\pdfoutput=1

\documentclass[11pt]{article}

\usepackage[]{acl}

\usepackage{times}
\usepackage{latexsym}
\usepackage{array}
\usepackage{multirow}
\usepackage{booktabs}

\usepackage{subcaption}

\usepackage[T1]{fontenc}

\usepackage[utf8]{inputenc}

\usepackage{microtype}

\usepackage{inconsolata}

\usepackage{graphicx} 

\title{Lost in the Middle, and In-Between: Enhancing Language Models' Ability to Reason Over Long Contexts in Multi-Hop QA}

\author{
\textnormal{$^\nabla$George Arthur Baker, Ankush Raut, $^\nabla$Sagi Shaier} \\
\textnormal{$^\dag$ Lawrence E Hunter, Katharina von der Wense$^{\nabla\diamondsuit}$} \\
$^\nabla$University of Colorado Boulder \\
$^\dag$University of Chicago, Department of Pediatrics\\
$^\diamondsuit$Johannes Gutenberg University Mainz \\
E-mail: \{george.baker, sagi.shaier, katharina.kann\}@colorado.edu}

\begin{document}
\maketitle
\begin{abstract}
Previous work finds that recent long-context language models fail to make equal use of information in the middle of their inputs, preferring pieces of information located at the tail ends which creates an undue bias in situations where we would like models to be equally capable of using different parts of the input. Thus far, the problem has mainly only been considered in settings with single pieces of critical information, leading us to question what happens when multiple necessary pieces of information are spread out over the inputs. Here, we demonstrate the effects of the "lost in the middle" problem in the multi-hop question answering  setting --- in which multiple reasoning "hops" over disconnected documents are required --- and show that performance degrades not only with respect to the distance of information from the edges of the context, but also between pieces of information. Additionally, we experiment with means of alleviating the problem by reducing superfluous document contents through knowledge graph triple extraction and summarization, and prompting models to reason more thoroughly using chain-of-thought prompting. We make our code and data available at: \href{https://github.com/Spongeorge/long-context-multihop}{https://github.com/Spongeorge/long-context-multihop}
\end{abstract}

\section{Introduction}

\begin{figure}[t]
    \centering
    \begin{subfigure}[t]{\linewidth}
        \centering
        \includegraphics{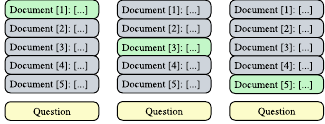}
        \caption{Previous work mainly explores the "lost in the middle" problem with single-hop questions, which contain just one gold document with the final answer, requiring minimal reasoning.}
        \label{fig:front_page_a}
    \end{subfigure}
    
    
    \begin{subfigure}[t]{\linewidth}
        \centering
        \includegraphics{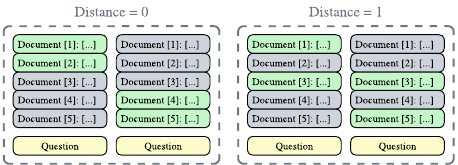}
        \caption{We explore models' performance with respect to the positions of multiple evidence documents, all of which must be reasoned over to answer a question correctly.}
         \label{fig:front_page_b}
    \end{subfigure}
    \caption{Question Answering setups with documents containing relevant information (green) and distractor documents (gray) placed at different ordinal positions, with both Single-hop (\ref{fig:front_page_a}) and Multi-hop (\ref{fig:front_page_b}) questions.}
    \label{fig:front_page}
\end{figure}

Recent advancements in attention mechanisms, such as Flash Attention \cite{dao2022flashattention, dao2023flashattention} and Attention with Linear Biases \cite{press2022train}, have ushered in a new generation of language models capable of handling significantly larger context sizes. These developments enable question-answering tasks to be performed over a substantial number of retrieved documents within a single input prompt (as shown in Figure \ref{fig:front_page}). However, despite this remarkable progress, recent studies reveal a critical limitation: long-context models fail to utilize information within their inputs equitably, exhibiting a pronounced bias toward information located at the edges of the context—a problem known as the "lost in the middle" \cite{liu2024lost}.

This limitation poses significant challenges for retrieval-augmented generation (RAG) systems, particularly in open-book question-answering scenarios. As the volume of input information increases, so does the likelihood that critical pieces of information necessary for answering a question may be overlooked \cite{shaier-etal-2024-adaptive}. This stands in contrast to earlier RAG systems, where documents were processed independently without explicit ordering, resulting in performance that typically improved with the inclusion of more retrieved documents \cite{DBLP:journals/corr/abs-2007-01282}.

Current efforts to address the "Lost in the Middle" problem primarily focus on approaches like document re-ranking \cite{peysakhovich2023attention, tang2023middle}, document length reduction via summarization \cite{kim2024sure}, or extending training to include long-context tasks \cite{an2024make}. However, these strategies face significant limitations in multi-hop QA settings. In such scenarios, where reasoning involves multiple steps across disconnected documents, the number of possible document order permutations grows combinatorially with the number of reasoning steps. This makes re-ranking and extended training approaches increasingly impractical as they would need to account for an overwhelming number of positional combinations. Similarly, the likelihood of omitting essential information through summarization grows with the number of reasoning steps, jeopardizing the integrity of reasoning chains.


In this work, we investigate the "Lost in the Middle" problem within the context of multi-hop QA. We argue that addressing this issue is critical for advancing the field, given the unique challenges it presents to current mitigation strategies. Specifically, we experiment with chain-of-thought (CoT) prompting \cite{zhou2023leasttomost} and document-size reduction methods to tackle this problem, utilizing recent long-context models such as GPT-3.5-Turbo, MPT-7b-instruct, and Llama-2-7b-longlora. Our key findings include:

\begin{enumerate}
    \item Performance degradation is not only influenced by the absolute position of information within the context but also by the relative distance between multiple relevant documents.
    \item Chain-of-Thought prompting aids in identifying relevant documents but fails to resolve the performance disparity caused by evidence document positions.
    \item Existing context reduction methods often produce reasoning chains that are too fragile for effective application in multi-hop QA.
\end{enumerate}

By highlighting these challenges and evaluating potential solutions, this study aims to guide future research in overcoming the "Lost in the Middle" problem and improving long-context model performance in complex multi-hop reasoning tasks.


\section{Related Work}
\subsection{The "Lost in the Middle" Problem}
The "Lost in the Middle" problem, first identified by \citet{liu2024lost}, highlights a significant limitation in long-context LMs. Specifically, when relevant information is distributed throughout a long context, model performance varies depending on the information's position. Their findings reveal that performance follows a characteristic curve: accuracy is poorest when critical information appears in the middle of the context and improves when the information is near the beginning or end.  


In their experiments with the NaturalQuestions-Open dataset \cite{kwiatkowski-etal-2019-natural}, \citet{liu2024lost} tested question-answering accuracy by repositioning the document containing the answer among distractor documents. They observed consistent variations in accuracy based on the document's position. Additionally, the authors studied a long-context key-value retrieval task, where models were tasked with retrieving a specific value associated with a key in an extended JSON file. Across both tasks, their results demonstrated that no examined model could process relevant information equally well across all positions.  


\subsection{Mitigation Strategies for "Lost in the Middle"}  
To mitigate the "Lost in the Middle" problem, \citet{liu2024lost} proposed Query-Aware Contextualization, which involves placing a query both before and after the request. While this approach effectively resolves the issue in the key-value retrieval setting, it has little impact on multi-document question answering, leaving the problem unresolved in that domain. 


Other efforts have focused on re-ranking passages before including them in the input prompt. \citet{peysakhovich2023attention} observed that LLMs assign preferential attention to relevant documents compared to irrelevant ones, even when located at the same position. They proposed sorting documents based on average attention scores while accounting for the typical attention distribution associated with positional biases.  


Similarly, \citet{tang2023middle} introduced "permutation self-consistency," a method that shuffles document orders and asks the model to rank their relevance, using a cumulative vote to determine the final order. However, these approaches are likely to scale poorly in the multi-hop QA setting. In multi-hop reasoning, later documents in the chain depend on earlier ones, making the number of permutations required for a robust ranking grow combinatorially with the number of reasoning steps.  


\subsection{Multi-Hop Question Answering}  


Multi-Hop Question Answering (MHQA) tasks involve reasoning across multiple documents \cite{yang-etal-2018-hotpotqa, saxena-etal-2020-improving, mavi2024multihopquestionanswering}, where relevant information is often distributed across the context in disconnected pieces. This often require models to combine parametric knowledge \cite{guo2022counterfactualmultihopqacauseeffect, feng-etal-2023-factkb, shaier-etal-2024-comparing, shaier-etal-2023-emerging, trivedi-etal-2020-multihop, su2024semistructuredchainofthoughtintegratingmultiple, lee2021robustifyingmultihopqapseudoevidentiality} with complex external context to derive answers. Unlike simpler QA tasks where all relevant information is co-located, the distributed nature of multihop reasoning poses significant challenges, often resulting in degraded performance.

The complexity of traversing reasoning chains across multiple sources not only impacts factuality \cite{guo2022counterfactualmultihopqacauseeffect, pezeshkpour2023measuringmodifyingfactualknowledge, wang2023surveyfactualitylargelanguage, shaier-etal-2023-stochastic, wang-etal-2024-factuality, su2024semistructuredchainofthoughtintegratingmultiple} but also hinders the model’s ability to consistently utilize all relevant information \cite{ yang-etal-2018-hotpotqa, su2024semistructuredchainofthoughtintegratingmultiple, shaier-etal-2024-desiderata, shaier-etal-2024-say, trivedi-etal-2020-multihop}. As input size and reasoning steps grow, maintaining factual accuracy becomes increasingly difficult, further compounding these challenges. 

\subsection{Impact of Input Length on Reasoning}  Simultaneously to our work, \citet{levy2024tokens} examined how LLM performance degrades with increasing input length. They found that reasoning performance deteriorates as the number of input tokens grows.  


Our study differs from theirs in several key aspects:  
\begin{enumerate}  
    \item We focus on performance with respect to document \textit{position} within a fixed-size context, whereas \citet{levy2024tokens} investigate performance relative to overall input size.  
    \item We evaluate on three popular multi-hop QA datasets, while \citet{levy2024tokens} use their custom dataset, FLenQA, which consists exclusively of true/false questions.  
    \item We study questions requiring up to four reasoning steps, whereas \citet{levy2024tokens} limit their analysis to two-step comparison questions.  
\end{enumerate}  
These distinctions emphasize our focus on understanding how positional biases within fixed contexts impact reasoning, particularly in multi-hop QA tasks.  




\section{Experiments}

\begin{table}
    \centering
    \begin{tabular}{lll}
    \hline
    \textbf{Dataset} & \textbf{Hops} & \textbf{Questions} \\
    \hline
    HotpotQA& 2 & 3703 \\
    2WikiMultihopQA\footnotemark& 2, 4 & 6288 \\
    MuSiQue-Ans& 2, 3, 4 & 1209 \\\hline
    \end{tabular}
    \caption{Multi-hop datasets we use to evaluate the Lost in the Middle problem. We use the 2nd half of the validation sets due to private test set labels.}
    \label{tab:datasets}
\end{table}

\footnotetext{As 2WikiMultihopQA contains only 10 documents per question, we retrieve an additional 10 distractor documents using a contriever setup as in \citet{liu2024lost}.}

\subsection{Datasets}
To evaluate language models on the "Lost in the Middle" problem in the multi-hop setting, we utilize existing Multi-hop Question Answering datasets (Table \ref{tab:datasets}). These datasets allow us to systematically position documents containing relevant information at various locations within the context, interspersed with distractor documents, to analyze how positional biases affect model performance.  


Since the official test sets for all three datasets are private and reserved for leaderboard purposes, we create our own splits by dividing the existing validation sets in half. The first half serves as our validation data, while the second half is used as our test set for reporting results.  



\subsubsection{Models}
To investigate the effects of the distance and position of evidence documents within a context on long-context language models, we experiment with a combination of popular open-source and closed-source models. Specifically, we use:  
\begin{itemize}  
    \item \textbf{MPT-7b-8k-instruct:} An instruction-tuned model trained with ALiBi \cite{press2022train}, which replaces traditional positional embeddings.  
    \item \textbf{Llama-2-7b-longlora-8k-ft} \cite{chen2023longlora}: A fine-tuned version of Llama 2 \cite{touvron2023llama} designed to support long contexts, without instruction tuning.  
    \item \textbf{GPT-3.5-turbo-1106:} One of the latest versions of OpenAI's GPT-3.5 Turbo, offering a context window of 16k tokens and newly reproducible outputs.  
\end{itemize}  

These models, representing a mix of open- and closed-source architectures, allow us to assess the generalizability of our findings across different LLMs.


\subsubsection{Metrics}
Following \citet{liu2024lost}, \citet{kandpal2023large}, and \citet{mallen2022not}, we adopt best-subspan accuracy as our evaluation metric. This metric assigns a score of 1 if the model's output contains the annotated answer (or any of the alternative answers, in the case of the MuSiQue dataset), and 0 otherwise.  


\subsubsection{Context Reduction Methods}
To investigate the relationship between context size and the "Lost in the Middle" problem, we extend our evaluation by applying two document size reduction methods:  

\begin{enumerate}  
    \item \textbf{Knowledge Graph Triple Extraction:} A technique that condenses documents into structured triples, capturing key facts while minimizing extraneous information.  
    \item \textbf{Document Summarization:} A method that generates concise summaries of documents, preserving their core content while reducing their overall length.  
\end{enumerate}  

By incorporating these reduction techniques, we aim to explore how modifying context size impacts the manifestation of the "Lost in the Middle" problem.  

\textbf{Knowledge Graph Triple Extraction}  
For extracting knowledge graph triples, we employ an instruction-based approach using the 7-billion-parameter version of LLaMA 2 \cite{touvron2023llama}. The model is prompted to extract triples from each individual document in the QA datasets, with the aim of capturing key factual relationships in a structured format.  

\textbf{Summarization}  
To summarize the documents in our datasets, we use BART-large-CNN \cite{DBLP:journals/corr/abs-1910-13461}, a pre-trained sequence-to-sequence model fine-tuned on the CNN/Daily Mail news summarization dataset \cite{hermann2015teaching}. Each document is processed independently, with the maximum generation length capped at 50 tokens to ensure concise summaries.


\begin{table}[]
    \centering
    \begin{tabular}{llll}
    \hline
    \textbf{Dataset} & \textbf{Full} & \textbf{Summ.} & \textbf{KG}  \\
    \hline
    HotpotQA & 69 & 29 & 33 \\
    2WikiMultihopQA & 45 & 21 & 29  \\
    MuSiQue-Ans & 85 & 32 & 35 \\\hline
    \end{tabular}
    \caption{Average document-wise word counts for each dataset and context-reduction method we use.}
    \label{tab:context_reduction_methods}
\end{table}

\begin{figure*}[p]
    \centering
    \begin{subfigure}{\textwidth}
        \centering        \includegraphics[width=\textwidth]{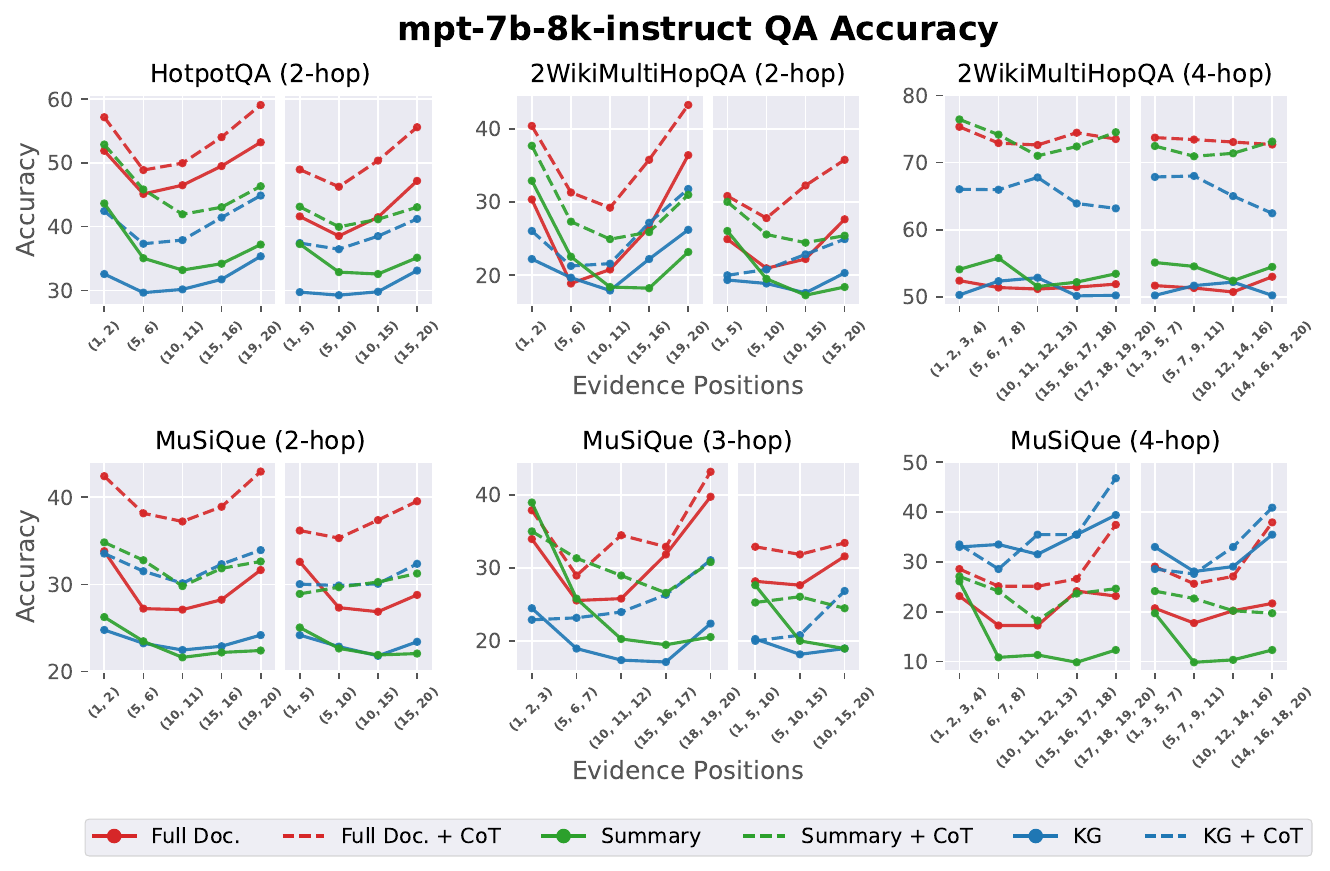}
        \label{fig:mpt_results_full}
    \end{subfigure}
     
    \begin{subfigure}{\textwidth}
        \centering
        \includegraphics[width=\textwidth]{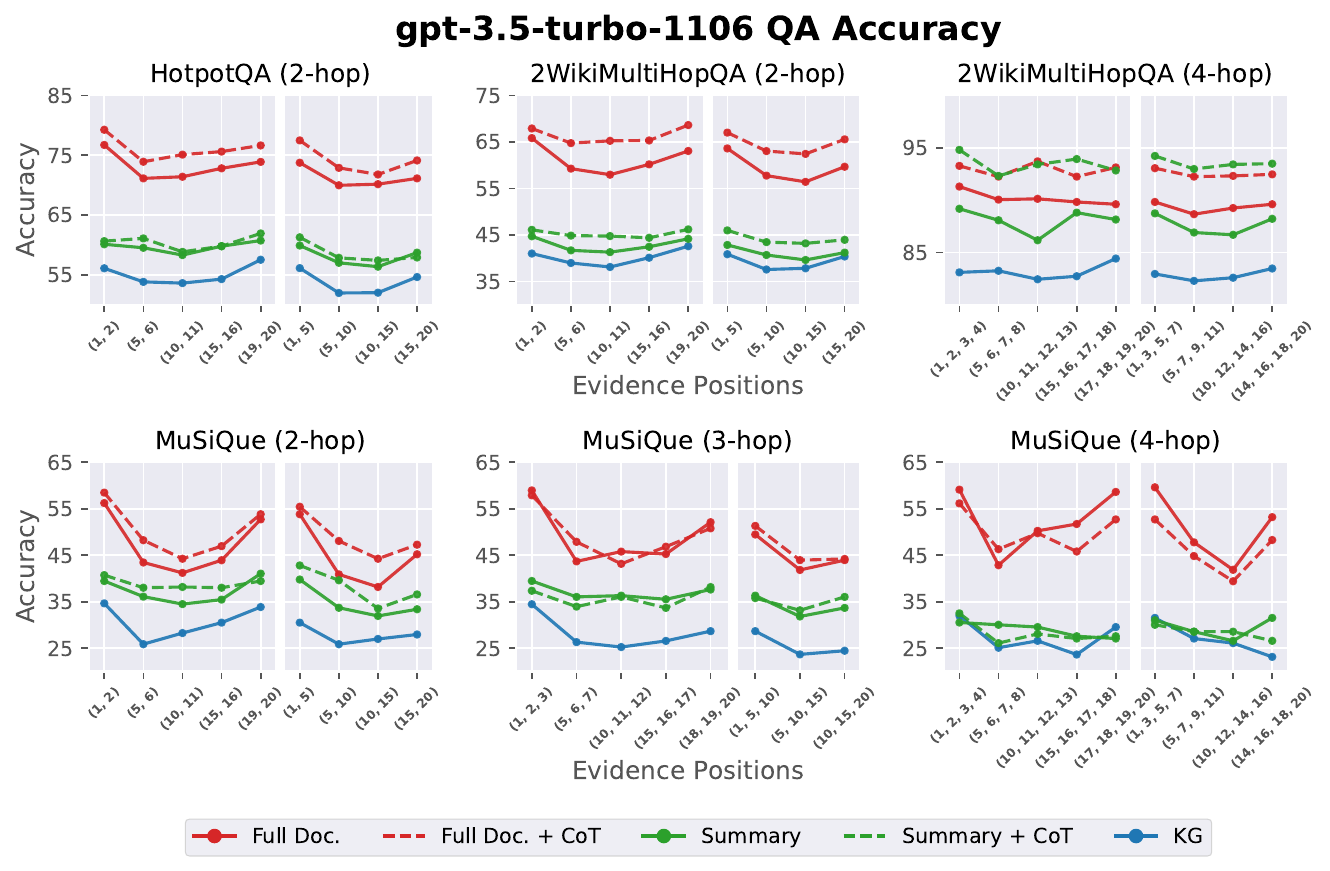}
         \label{fig:gpt_results_full}
    \end{subfigure}

    \caption{The performance impacts of varying the positions of relevant documents within instruction-tuned models' inputs, with context reduction techniques and Chain-of-Thought prompting. All positions are out of 20 total documents. KG + CoT results for gpt-3.5-turbo are omitted to Appendix \ref{sec:full_results} to highlight other results.}
    \label{fig:results_main}
\end{figure*}

\begin{figure}[]
\includegraphics[width=.9\linewidth]{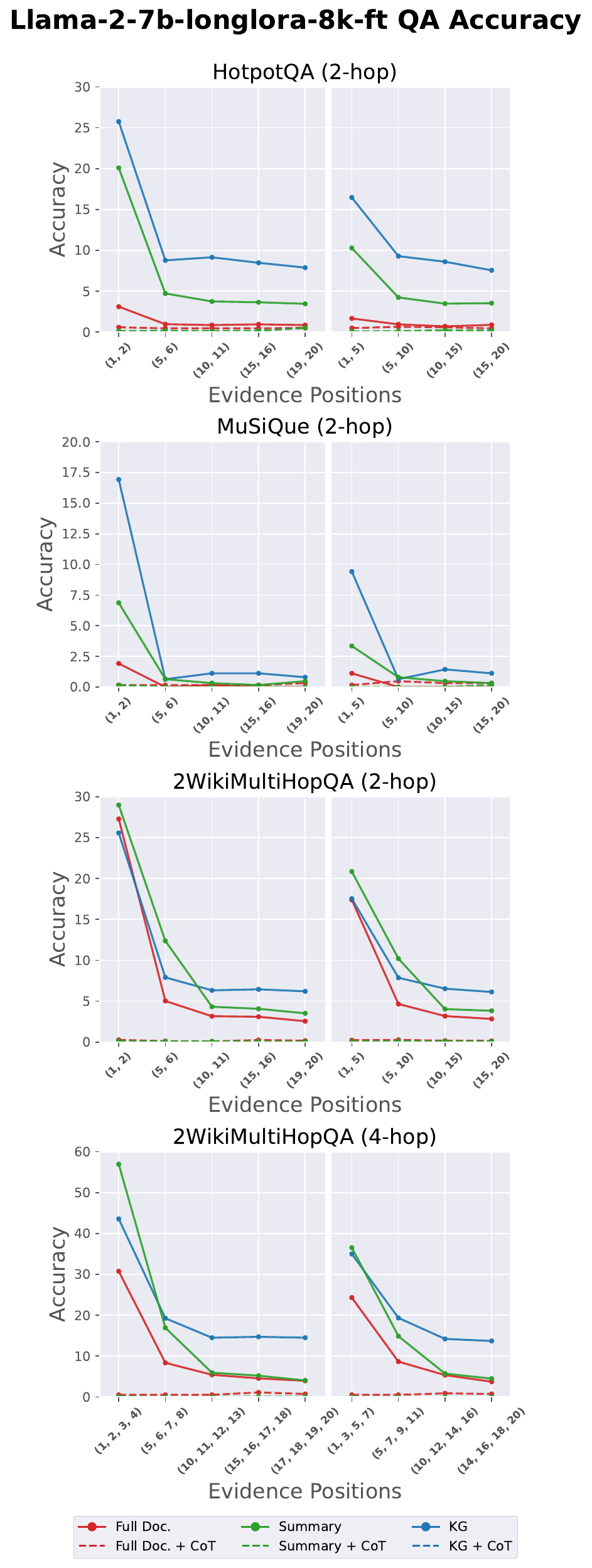}
    \caption{Experimental results for Llama-2-7b-longlora-8k-ft. Results for MuSiQue 3- and 4-hop splits are relegated to Appendix \ref{sec:full_results} due to exceedingly poor performance.}
    \label{fig:non_instruct_results}
\end{figure}

\begin{figure*}[]
    \centering
    \begin{subfigure}[t]{\linewidth}
        \centering
        \includegraphics[width=\linewidth]{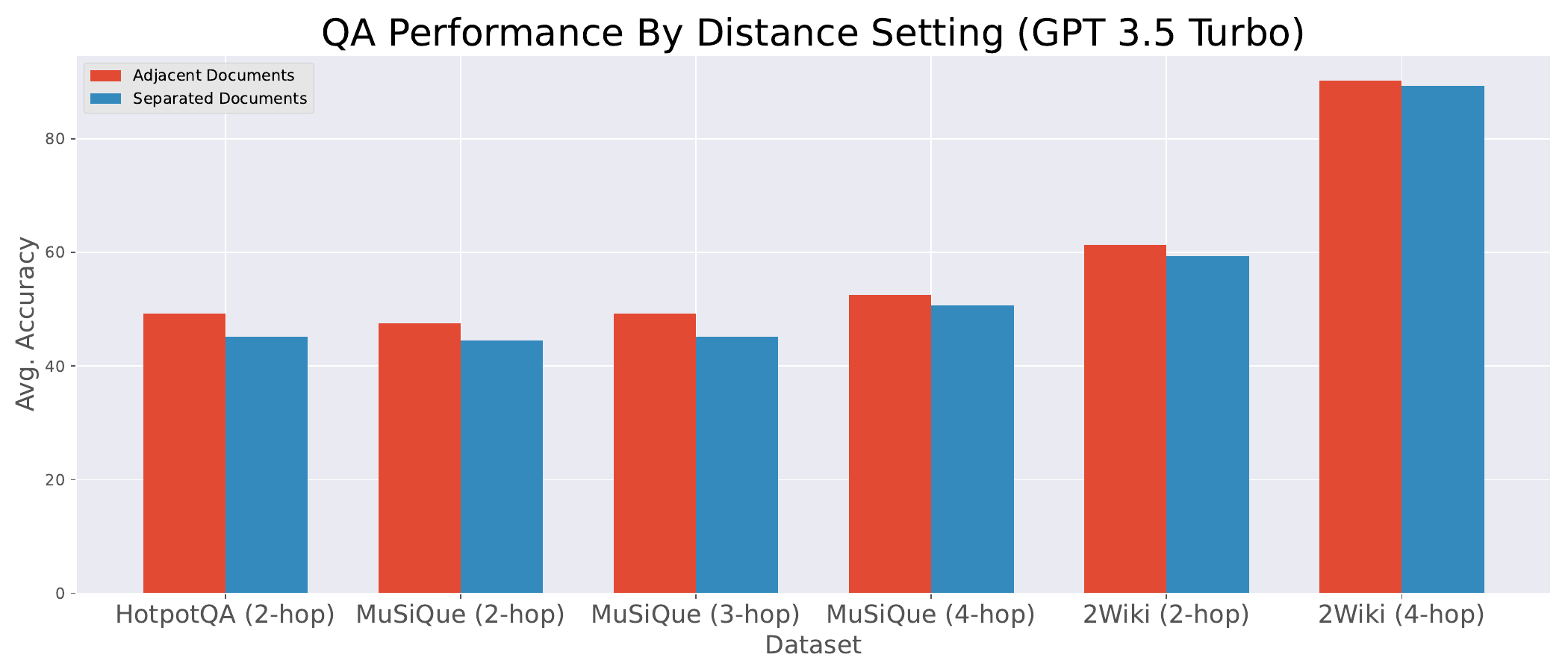}
        \caption{Average question-answering accuracy for GPT models with full document prompts by distance setting. Performance is generally higher when evidence documents are adjacent compared to when they are separated by distractor documents.}
        \label{fig:gpt_perf_by_dist}
    \end{subfigure}
    \vspace{0.5cm} 
    \begin{subfigure}[t]{\linewidth}
        \centering
        \includegraphics[width=\linewidth]{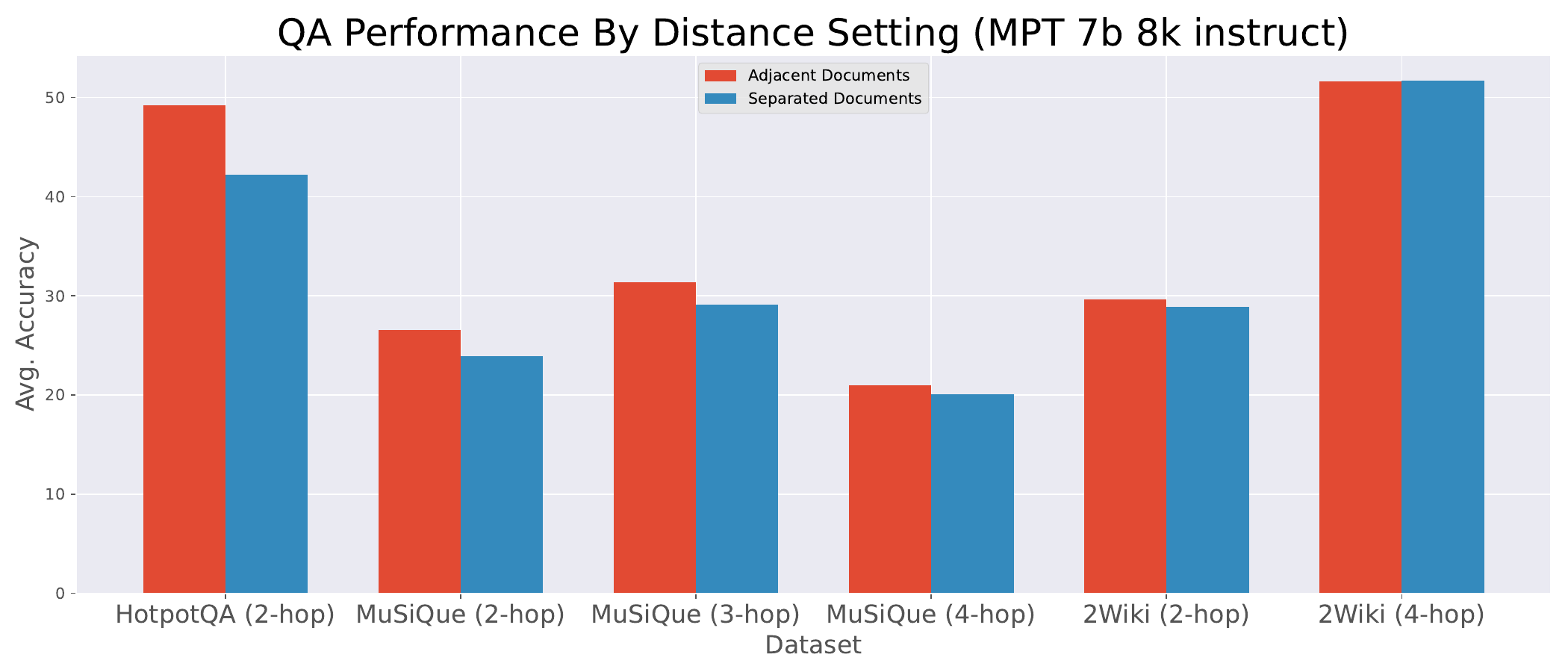}
        \caption{Average question-answering accuracy for MPT models with full document prompts by distance setting. Similar to GPT, performance decreases as the separation between evidence documents increases.}
        \label{fig:mpt_perf_by_dist}
    \end{subfigure}
    \caption{Average question-answering accuracy for full document prompts by distance setting for GPT and MPT models. Performance with adjacent evidence documents is generally higher than when evidence documents are separated by distractor documents.}
    \label{fig:analysis}
\end{figure*}

\section{Methodology}
For each question in the datasets (and their context-reduced variants), we create multiple prompts by positioning evidence documents at various locations within a total of 20 documents, following the approach outlined by \citet{liu2024lost}. Given the high number of potential combinations of evidence positions (${n \choose k} = 190, 1140, 4845$ possible orderings per prompt in our 2-, 3-, and 4-hop settings, where $n$ represents the number of document positions and $k$ the number of evidence documents), we select a subset of combinations to manage the computational cost of generating long-context prompts. Specifically, we choose 5 combinations where the gold documents are placed adjacent to one another, as well as 4 combinations with distractor documents interspersed between the gold documents for 2- and 4-hop questions, and 3 combinations for 3-hop questions. The distractor documents are retained in their original order, based on their relevance determined through the retrieval process.

The prompts are processed by the models with a temperature setting of 0, and the generation is limited to a maximum of 256 tokens.



\section{Results}
The results for instruction-tuned models are presented in Figure \ref{fig:results_main}, while those for non-instruction-tuned models are shown in Figure \ref{fig:non_instruct_results}. For a comprehensive view of all results, including instances where models performed suboptimally, please refer to Appendix \ref{sec:full_results}. We focus here on the most notable findings.

Specifically, Figures \ref{fig:results_main} and \ref{fig:non_instruct_results} illustrate the performance variations across the three models, with respect to each dataset, as we manipulate the placement of the relevant documents (e.g., positions 1 and 2, 5 and 6, etc.). Additionally, the figures compare model performance when using the full document, a summarized version, or knowledge graph triples extracted from the document. For each condition, we experiment with and without CoT prompting to assess its impact on model performance.

Lastly, Figure \ref{fig:analysis} provides a detailed analysis of how the distance between relevant documents impacts model performance across different datasets. Specifically, it examines the performance variations when relevant documents are placed adjacently at different positions, compared to scenarios where they are separated by distractor documents. This analysis highlights the sensitivity of the models to the spatial arrangement of evidence and underscores the challenges posed by non-adjacent evidence placement in the context of multi-hop reasoning tasks.




\section{Analysis}
\textbf{Proximity of relevant documents significantly affects performance.}
Figure~\ref{fig:analysis} highlights a clear trend: models perform better when relevant documents are adjacent compared to when they are separated by distractor documents. This suggests that the spatial proximity of evidence is crucial for models to effectively retrieve and integrate information. When documents are adjacent, connections between them are more easily captured, potentially due to the model's attention mechanisms. In contrast, when distractors separate the relevant documents, the models struggle to retrieve and synthesize the necessary evidence, resulting in a drop in performance.


\textbf{Chain-of-Thought prompting yields mixed results.}
For instruction-tuned models such as MPT and GPT-3.5, CoT prompting with few-shot exemplars markedly improves performance compared to zero-shot settings in most scenarios. This improvement likely stems from the explicit reasoning steps provided by CoT, which help these models better structure their responses and navigate complex multi-hop reasoning tasks. However, for the non-instruction-tuned Llama 2 longlora model, CoT prompting leads to a sharp decline in performance. This discrepancy may stem from the model's inherent biases: it demonstrates a pronounced primacy bias, with little to no recency bias, leading to an over-reliance on the few-shot exemplars and improper integration of the actual task-specific context. This suggests that non-instruction-tuned models may require careful tuning or additional training to fully leverage CoT-style reasoning.

\textbf{Context reduction mitigates position biases but sacrifices accuracy.}
Figure~\ref{fig:results_main} reveals that reducing context—whether through summarization or knowledge graph triple extraction—dampens the impact of the "lost in the middle" problem. Specifically, evidence located in the middle of the input achieves performance levels closer to those of edge-positioned evidence when using reduced contexts. This flattening of the performance curve suggests that context reduction alleviates the models' positional biases. However, this improvement comes at a cost: overall accuracy declines, particularly in instruction-tuned models like MPT and GPT-3.5. This drop is likely due to information loss during the context reduction process. Interestingly, the non-instruction-tuned Llama 2 longlora model benefits substantially from context reduction, suggesting that these methods can serve as a useful preprocessing step for less robust models.


\section{Future Work}

Building on the findings of this study, there are several promising directions for future research aimed at addressing the "Lost in the Middle" problem in multi-hop reasoning tasks.

First, while our work highlights the impact of document positioning and adjacency, a more exhaustive evaluation of evidence combinations is needed. Computational constraints limited our analysis to a subset of possible configurations, leaving unexplored how more complex or nuanced arrangements affect model performance. Advances in more efficient evaluation methodologies or sampling strategies could enable a broader and deeper exploration of this space.

Second, the limitations of current context-reduction techniques, such as summarization and knowledge graph triple extraction, underscore the need for improved preprocessing methods. Future research could develop tailored strategies that preserve key reasoning paths while minimizing extraneous information. Techniques leveraging recent advancements in unsupervised learning, retrieval-augmented generation, or specialized fine-tuning could prove effective in mitigating information loss.

Third, the disparity in performance improvements between instruction-tuned and non-instruction-tuned models suggests an opportunity to better align model architectures with task-specific reasoning demands. Investigating the potential of advanced prompting techniques, such as dynamic CoT, could enhance model performance, particularly for non-instruction-tuned models like Llama 2 longlora.

Additionally, exploring the robustness of newer and larger models to the "Lost in the Middle" problem remains an open question. Recent advancements in large-scale pretraining and reasoning capabilities may yield models better equipped to handle dispersed evidence, and future evaluations should incorporate these state-of-the-art systems to identify potential improvements.

Finally, incorporating external memory mechanisms or augmenting model architectures to dynamically prioritize and retrieve relevant evidence could offer a path forward. Such modifications may allow models to better handle long-context reasoning challenges, reducing sensitivity to document positioning and improving overall robustness in multi-hop settings.

By addressing these directions, future research can move closer to resolving the challenges posed by dispersed evidence in multi-hop question answering and enhancing the capabilities of long-context language models in reasoning-intensive tasks.

\section{Conclusion}
In this paper, we presented a study of the "Lost in the Middle" problem in the context of Multi-hop Question Answering, where models must integrate information from multiple documents to generate correct answers. Using three widely-used multi-hop QA datasets (HotpotQA, 2WikiMultihopQA, and MuSiQue-Ans), we analyzed the performance of recent large language models as a function of the positions of evidence documents within a context interspersed with distractors. Our findings reveal that model performance is not only influenced by the absolute positions of evidence documents but also by their relative positioning, highlighting a previously underexplored dimension of this problem.

We also explored context-reduction techniques such as summarization and knowledge graph triple extraction as potential solutions. However, our results indicate that these out-of-the-box approaches are insufficient to fully mitigate the issue, as they often lead to a trade-off between reducing positional bias and retaining critical information.

Overall, our work underscores the complexity of the "Lost in the Middle" problem in multi-hop settings, extending beyond the single-hop scenarios that current mitigation strategies typically address. This study opens up new challenges for the design of LLMs and methods aimed at improving reasoning across long and complex contexts.


\section*{Limitations}
Our evaluation of the "Lost in the Middle" problem in multi-hop settings is subject to several limitations, primarily due to computational and time constraints. First, we assessed a carefully curated subset of possible evidence location combinations rather than exhaustively evaluating all permutations. The factorial growth in combinations, combined with the high token counts of our prompts, made a comprehensive analysis computationally prohibitive. While our approach highlights meaningful patterns, it may not capture the full extent of potential variations in performance.

Second, although the models evaluated were all state-of-the-art at the time of this study and released within the past year, newer and potentially more powerful models have since become available. These more recent models, with greater reasoning capabilities and larger parameter counts, may exhibit improved robustness against the "Lost in the Middle" problem. Investigating the behavior of these models remains an important direction for future work.

Lastly, our experiments primarily focus on out-of-the-box summarization and knowledge graph extraction techniques for context reduction. More advanced, model-specific fine-tuning or optimized preprocessing strategies may yield better results, but these were beyond the scope of our current study. Future research could explore such targeted interventions to further address the challenges identified here.

\bibliography{custom}

\appendix

\section{Full Results}
\label{sec:full_results}

\begin{table*}[htbp]
\centering
\tiny
\begin{tabular}{|l|l|c|c|c|c|c|c|c|c|c|}
\hline
Dataset (Closed-book Score) & Prompt & 1,2 & 5,6 & 10,11 & 15,16 & 19,20 & 1,5 & 5,10 & 10,15 & 15,20 \\ \hline
\multirow{6}{*}{Hotpot (40.64\%)} & standard & 76.72\% & 71.13\% & 71.40\% & 72.81\% & 73.89\% & 73.75\% & 69.97\% & 70.16\% & 71.13\% \\
& standard + CoT & 79.26\% & 73.91\% & 75.10\% & 75.61\% & 76.64\% & 77.48\% & 72.89\% & 71.78\% & 74.13\% \\
& kg & 56.09\% & 53.82\% & 53.61\% & 54.28\% & 57.52\% & 56.12\% & 51.96\% & 52.01\% & 54.63\% \\
& kg + CoT & 34.94\% & 28.03\% & 29.57\% & 29.92\% & 31.14\% & 31.60\% & 28.25\% & 28.54\% & 29.84\% \\
& summaries & 60.11\% & 59.52\% & 58.30\% & 59.79\% & 60.73\% & 59.90\% & 57.01\% & 56.36\% & 58.71\% \\
& summaries + CoT & 60.63\% & 61.09\% & 58.84\% & 59.79\% & 61.92\% & 61.27\% & 57.84\% & 57.41\% & 57.90\% \\ \hline

\multirow{1}{*}{ MuSiQue (14.39\%)} & \phantom{} & 1,2 & 5,6 & 10,11 & 15,16 & 19,20 & 1,5 & 5,10 & 10,15 & 15,20 \\ \hline

\multirow{6}{*}{2-hop} & standard & 56.23\% & 43.45\% & 41.21\% & 43.93\% & 52.72\% & 53.83\% & 40.89\% & 38.18\% & 45.21\% \\
& standard + CoT & 58.47\% & 48.24\% & 44.25\% & 46.96\% & 53.83\% & 55.43\% & 48.08\% & 44.25\% & 47.28\% \\
& kg & 34.66\% & 25.88\% & 28.27\% & 30.51\% & 33.87\% & 30.51\% & 25.88\% & 27.00\% & 27.96\% \\
& kg + CoT & 14.06\% & 7.67\% & 9.90\% & 9.11\% & 10.86\% & 12.14\% & 8.63\% & 9.11\% & 9.90\% \\
& summaries & 39.46\% & 36.10\% & 34.50\% & 35.46\% & 41.05\% & 39.78\% & 33.71\% & 31.95\% & 33.39\% \\
& summaries + CoT & 40.73\% & 38.02\% & 38.18\% & 38.02\% & 39.46\% & 42.81\% & 39.62\% & 33.55\% & 36.58\% \\ \hline

\multirow{1}{*}{}  &  & 1, 2, 3 & 5, 6, 7 & 10, 11, 12 & 15, 16, 17 & 18, 19,20 & 1, 5, 10 & 5, 10, 15 & 10, 15, 20 &  \\ \hline

\multirow{6}{*}{3-hop} & standard & 58.95\% & 43.68\% & 45.79\% & 45.26\% & 52.11\% & 49.47\% & 41.84\% & 43.95\% & - \\
& standard + CoT & 57.89\% & 47.89\% & 43.16\% & 46.84\% & 50.79\% & 51.32\% & 43.95\% & 44.21\% & - \\
& kg & 34.47\% & 26.32\% & 25.26\% & 26.58\% & 28.68\% & 28.68\% & 23.68\% & 24.47\% & - \\
& kg + CoT & 5.79\% & 3.68\% & 5.79\% & 5.79\% & 5.79\% & 5.53\% & 5.53\% & 4.47\% & - \\
& summaries & 39.47\% & 36.05\% & 36.32\% & 35.53\% & 37.63\% & 36.32\% & 31.84\% & 33.68\% & - \\
& summaries + CoT & 37.37\% & 33.95\% & 36.05\% & 33.68\% & 38.16\% & 35.79\% & 33.16\% & 36.05\% & - \\ \hline

\multirow{1}{*}{}  & \phantom{} & 1,2,3,4 & 5,6,7,8 & 10,11,12,13 & 15,16,17,18 & 17,18, 19,20 & 1,3,5,7 & 5,7,9,11 & 10,12,14,16 & 14,16,18,20 \\ \hline

\multirow{6}{*}{4-hop} & standard & 59.11\% & 42.86\% & 50.25\% & 51.72\% & 58.62\% & 59.61\% & 47.78\% & 41.87\% & 53.20\% \\
& standard + CoT & 56.16\% & 46.31\% & 49.75\% & 45.81\% & 52.71\% & 52.71\% & 44.83\% & 39.41\% & 48.28\% \\
& kg & 32.02\% & 25.12\% & 26.60\% & 23.65\% & 29.56\% & 31.53\% & 27.09\% & 26.11\% & 23.15\% \\
& kg + CoT & 3.94\% & 3.45\% & 5.91\% & 5.91\% & 5.91\% & 4.93\% & 5.91\% & 3.94\% & 5.91\% \\
& summaries & 30.54\% & 30.05\% & 29.56\% & 27.59\% & 27.09\% & 31.03\% & 28.57\% & 26.60\% & 31.53\% \\
& summaries + CoT & 32.51\% & 26.11\% & 28.08\% & 27.09\% & 27.59\% & 30.05\% & 28.57\% & 28.57\% & 26.60\% \\ \hline

\multirow{1}{*}{2Wiki (44.99\%)} &  & 1,2 & 5,6 & 10,11 & 15,16 & 19,20 & 1,5 & 5,10 & 10,15 & 15,20 \\ \hline

\multirow{6}{*}{2-hop} & standard & 65.83\% & 59.26\% & 57.96\% & 60.18\% & 63.06\% & 63.61\% & 57.76\% & 56.42\% & 59.67\% \\
& standard + CoT & 67.90\% & 64.75\% & 65.24\% & 65.32\% & 68.63\% & 67.01\% & 63.04\% & 62.43\% & 65.56\% \\
& kg & 41.00\% & 38.97\% & 38.11\% & 40.11\% & 42.58\% & 40.86\% & 37.57\% & 37.83\% & 40.37\% \\
& kg + CoT & 25.99\% & 24.08\% & 24.32\% & 24.46\% & 26.11\% & 25.05\% & 24.38\% & 23.77\% & 24.42\% \\
& summaries & 44.72\% & 41.69\% & 41.30\% & 42.46\% & 44.17\% & 42.85\% & 40.69\% & 39.62\% & 41.20\% \\
& summaries + CoT & 46.10\% & 44.86\% & 44.76\% & 44.39\% & 46.20\% & 45.98\% & 43.46\% & 43.19\% & 43.95\% \\ \hline

\multirow{1}{*}{}  & \phantom{} & 1,2,3,4 & 5,6,7,8 & 10,11,12,13 & 15,16,17,18 & 17,18, 19,20 & 1,3,5,7 & 5,7,9,11 & 10,12,14,16 & 14,16,18,20 \\ \hline

\multirow{6}{*}{4-hop} & standard & 91.29\% & 90.04\% & 90.12\% & 89.82\% & 89.60\% & 89.82\% & 88.65\% & 89.24\% & 89.60\% \\
& standard + CoT & 93.27\% & 92.24\% & 93.70\% & 92.24\% & 93.12\% & 93.05\% & 92.24\% & 92.31\% & 92.46\% \\
& kg & 83.09\% & 83.24\% & 82.43\% & 82.72\% & 84.41\% & 82.94\% & 82.28\% & 82.58\% & 83.46\% \\
& kg + CoT & 65.15\% & 67.20\% & 67.94\% & 64.28\% & 64.13\% & 66.62\% & 65.96\% & 65.81\% & 63.47\% \\
& summaries & 89.17\% & 88.07\% & 86.16\% & 88.80\% & 88.14\% & 88.73\% & 86.90\% & 86.68\% & 88.21\% \\
& summaries + CoT & 94.80\% & 92.31\% & 93.41\% & 93.92\% & 92.83\% & 94.22\% & 92.97\% & 93.41\% & 93.48\% \\ \hline

\end{tabular}
\caption{Full experimental results of gpt-3.5-turbo-1106. Percentages next to dataset names are the closed-book scores for the full set.}
\label{tab:gpt_results}
\end{table*}

\begin{table*}[htbp]
\centering
\tiny
\begin{tabular}{|l|l|c|c|c|c|c|c|c|c|c|}
\hline
Dataset (Closed-book Score) & Prompt & 1,2 & 5,6 & 10,11 & 15,16 & 19,20 & 1,5 & 5,10 & 10,15 & 15,20 \\ \hline
\multirow{6}{*}{Hotpot (14.88\%)} & standard & 51.90\% & 45.13\% & 46.50\% & 49.50\% & 53.23\% & 41.61\% & 38.54\% & 41.48\% & 47.18\% \\
& standard + CoT & 57.17\% & 48.88\% & 49.96\% & 54.04\% & 59.09\% & 48.96\% & 46.26\% & 50.36\% & 55.60\% \\
& kg & 32.57\% & 29.65\% & 30.16\% & 31.73\% & 35.35\% & 29.73\% & 29.25\% & 29.79\% & 33.11\% \\
& kg + CoT & 42.45\% & 37.32\% & 37.89\% & 41.43\% & 44.88\% & 37.43\% & 36.46\% & 38.51\% & 41.21\% \\
& summaries & 43.64\% & 35.05\% & 33.19\% & 34.19\% & 37.19\% & 37.27\% & 32.87\% & 32.57\% & 35.13\% \\
& summaries + CoT & 52.88\% & 45.83\% & 41.94\% & 43.05\% & 46.34\% & 43.13\% & 39.97\% & 41.16\% & 43.05\% \\ \hline

\multirow{6}{*}{}Musique (3.64\%) &  & 1,2 & 5,6 & 10,11 & 15,16 & 19,20 & 1,5 & 5,10 & 10,15 & 15,20 \\ \hline
2-hop & standard & 30.35\% & 18.85\% & 20.77\% & 26.52\% & 36.42\% & 24.92\% & 20.93\% & 22.20\% & 27.64\% \\
& standard + CoT & 40.42\% & 31.31\% & 29.23\% & 35.78\% & 43.29\% & 30.83\% & 27.80\% & 32.27\% & 35.78\% \\
& kg & 22.20\% & 19.65\% & 17.89\% & 22.20\% & 26.20\% & 19.33\% & 18.85\% & 17.57\% & 20.29\% \\
& kg + CoT & 26.04\% & 21.25\% & 21.57\% & 27.16\% & 31.79\% & 19.97\% & 20.77\% & 22.84\% & 24.92\% \\
& summaries & 32.91\% & 22.52\% & 18.37\% & 18.21\% & 23.16\% & 26.04\% & 19.49\% & 17.25\% & 18.37\% \\
& summaries + CoT & 37.70\% & 27.32\% & 24.92\% & 25.88\% & 30.99\% & 30.03\% & 25.56\% & 24.44\% & 25.40\% \\ \hline

\multirow{1}{*}{}  &  & 1,2,3 & 5,6,7 & 10,11,12 & 15,16,17 & 18,19,20 & 1,5,10 & 5,10,15 & 10,15,20 &  \\ \hline

\multirow{6}{*}{3-hop} & standard & 33.95\% & 25.53\% & 25.79\% & 31.84\% & 39.74\% & 28.16\% & 27.63\% & 31.58\% & - \\ 
& standard + CoT & 37.89\% & 28.95\% & 34.47\% & 32.89\% & 43.16\% & 32.89\% & 31.84\% & 33.42\% & - \\ 
& kg & 24.47\% & 18.95\% & 17.37\% & 17.11\% & 22.37\% & 20.26\% & 18.16\% & 18.95\% & - \\ 
& kg + CoT & 22.89\% & 23.16\% & 23.95\% & 26.32\% & 31.05\% & 20.00\% & 20.79\% & 26.84\% & - \\ 
& summaries & 38.95\% & 25.79\% & 20.26\% & 19.47\% & 20.53\% & 27.63\% & 20.00\% & 18.95\% & - \\ 
& summaries + CoT & 35.00\% & 31.32\% & 28.95\% & 26.58\% & 30.79\% & 25.26\% & 26.05\% & 24.47\% & - \\ \hline

\multirow{1}{*}{}  &  & 1,2,3,4 & 5,6,7,8 & 10,11,12,13 & 15,16,17,18 & 17,18, 19,20 & 1,3,5,7 & 5,7,9,11 & 10,12,14,16 & 14,16,18,20 \\ \hline

\multirow{6}{*}{4-hop} & standard & 23.15\% & 17.24\% & 17.24\% & 24.14\% & 23.15\% & 20.69\% & 17.73\% & 20.20\% & 21.67\% \\
& standard + CoT & 28.57\% & 25.12\% & 25.12\% & 26.60\% & 37.44\% & 29.06\% & 25.62\% & 27.09\% & 37.93\% \\
& kg & 33.00\% & 33.50\% & 31.53\% & 35.47\% & 39.41\% & 33.00\% & 28.08\% & 29.06\% & 35.47\% \\
& kg + CoT & 33.50\% & 28.57\% & 35.47\% & 35.47\% & 46.80\% & 28.57\% & 27.59\% & 33.00\% & 40.89\% \\
& summaries & 26.11\% & 10.84\% & 11.33\% & 9.85\% & 12.32\% & 19.70\% & 9.85\% & 10.34\% & 12.32\% \\
& summaries + CoT & 27.09\% & 24.14\% & 18.23\% & 23.65\% & 24.63\% & 24.14\% & 22.66\% & 20.20\% & 19.70\% \\ \hline

\multirow{1}{*}{2wiki (20.13\%)} &  & 1,2 & 5,6 & 10,11 & 15,16 & 19,20 & 1,5 & 5,10 & 10,15 & 15,20 \\ \hline
\multirow{6}{*}{2-hop} & standard & 33.81\% & 27.22\% & 27.10\% & 28.24\% & 31.65\% & 32.59\% & 27.33\% & 26.86\% & 28.79\% \\
& standard + CoT & 42.42\% & 38.16\% & 37.22\% & 38.91\% & 42.95\% & 36.18\% & 35.31\% & 37.38\% & 39.54\% \\
& kg & 24.77\% & 23.24\% & 22.47\% & 22.90\% & 24.18\% & 24.18\% & 22.86\% & 21.80\% & 23.41\% \\
& kg + CoT & 33.54\% & 31.51\% & 30.13\% & 32.32\% & 33.93\% & 30.03\% & 29.87\% & 30.07\% & 32.36\% \\
& summaries & 26.25\% & 23.47\% & 21.62\% & 22.19\% & 22.41\% & 25.05\% & 22.65\% & 21.90\% & 22.06\% \\
& summaries + CoT & 34.82\% & 32.77\% & 29.80\% & 31.84\% & 32.63\% & 28.91\% & 29.68\% & 30.27\% & 31.25\% \\ \hline

\multirow{1}{*}{} &  & 1,2,3,4 & 5,6,7,8 & 10,11,12,13 & 15,16,17,18 & 17,18, 19,20 & 1,3,5,7 & 5,7,9,11 & 10,12,14,16 & 14,16,18,20 \\ \hline

\multirow{6}{*}{4-hop} & standard & 52.42\% & 51.39\% & 51.17\% & 51.46\% & 51.90\% & 51.68\% & 51.32\% & 50.73\% & 53.00\% \\
& standard + CoT & 75.33\% & 72.91\% & 72.62\% & 74.45\% & 73.50\% & 73.72\% & 73.43\% & 73.06\% & 72.69\% \\
& kg & 50.29\% & 52.34\% & 52.86\% & 50.15\% & 50.22\% & 50.22\% & 51.68\% & 52.20\% & 50.22\% \\
& kg + CoT & 66.03\% & 65.96\% & 67.79\% & 63.91\% & 63.18\% & 67.86\% & 68.01\% & 65.01\% & 62.45\% \\
& summaries & 54.10\% & 55.78\% & 51.54\% & 52.20\% & 53.44\% & 55.12\% & 54.54\% & 52.42\% & 54.47\% \\
& summaries + CoT & 76.43\% & 74.16\% & 71.01\% & 72.40\% & 74.52\% & 72.47\% & 70.94\% & 71.38\% & 73.13\% \\ \hline
\end{tabular}
\caption{Full experimental results of mpt-7b-8k-instruct. Percentages next to dataset names are the closed-book scores for the full set.}
\label{table:mpt_results}
\end{table*}

\begin{table*}[htbp]
\centering
\tiny
\begin{tabular}{|l|l|c|c|c|c|c|c|c|c|c|}
\hline
Dataset (Closed-book Score) & Prompt & 1,2 & 5,6 & 10,11 & 15,16 & 19,20 & 1,5 & 5,10 & 10,15 & 15,20 \\ \hline

\multirow{6}{*}{Hotpot (12.21\%)} & standard & 3.11\% & 0.97\% & 0.86\% & 0.95\% & 0.86\% & 1.67\% & 0.95\% & 0.70\% & 0.89\% \\
& standard + CoT & 0.59\% & 0.46\% & 0.46\% & 0.46\% & 0.51\% & 0.49\% & 0.65\% & 0.59\% & 0.46\% \\
& kg & 25.76\% & 8.78\% & 9.15\% & 8.48\% & 7.89\% & 16.47\% & 9.29\% & 8.61\% & 7.56\% \\
& kg + CoT & 1.16\% & 0.97\% & 1.03\% & 1.05\% & 0.81\% & 0.89\% & 1.08\% & 0.89\% & 1.08\% \\
& summaries & 20.09\% & 4.73\% & 3.75\% & 3.65\% & 3.46\% & 10.29\% & 4.24\% & 3.48\% & 3.54\% \\
& summaries + CoT & 0.14\% & 0.16\% & 0.11\% & 0.16\% & 0.51\% & 0.08\% & 0.11\% & 0.24\% & 0.19\% \\ \hline

\multirow{6}{*}{}Musique (0.74\%) &  & 1,2 & 5,6 & 10,11 & 15,16 & 19,20 & 1,5 & 5,10 & 10,15 & 15,20 \\ \hline
\multirow{6}{*}{2-hop} & standard & 1.92\% & 0.00\% & 0.16\% & 0.00\% & 0.00\% & 1.12\% & 0.00\% & 0.00\% & 0.00\% \\
& standard + CoT & 0.16\% & 0.16\% & 0.16\% & 0.16\% & 0.32\% & 0.16\% & 0.48\% & 0.32\% & 0.32\% \\
& kg & 16.93\% & 0.64\% & 1.12\% & 1.12\% & 0.80\% & 9.42\% & 0.64\% & 1.44\% & 1.12\% \\
& kg + CoT & 0.16\% & 0.00\% & 0.16\% & 0.16\% & 0.16\% & 0.16\% & 0.16\% & 0.16\% & 0.16\% \\
& summaries & 6.87\% & 0.64\% & 0.32\% & 0.16\% & 0.48\% & 3.35\% & 0.80\% & 0.48\% & 0.32\% \\
& summaries + CoT & 0.16\% & 0.00\% & 0.00\% & 0.00\% & 0.00\% & 0.00\% & 0.00\% & 0.00\% & 0.16\% \\ \hline

\multirow{6}{*}{} &  & 1,2,3 & 5,6,7 & 10,11,12 & 15,16,17 & 18,19,20 & 1,5,10 & 5,10,15 & 10,15,20 & \\ \hline
\multirow{6}{*}{3-hop} & standard & 0.00\% & 0.00\% & 0.00\% & 0.00\% & 0.00\% & 0.00\% & 0.00\% & 0.00\% & - \\
& standard + CoT & 0.00\% & 0.00\% & 0.26\% & 0.00\% & 0.00\% & 0.00\% & 0.26\% & 0.26\% & - \\
& kg & 0.00\% & 0.00\% & 0.00\% & 0.26\% & 0.26\% & 0.00\% & 0.00\% & 0.26\% & - \\
& kg + CoT & 0.26\% & 0.26\% & 0.26\% & 0.26\% & 0.00\% & 0.26\% & 0.26\% & 0.00\% & - \\
& summaries & 0.00\% & 0.00\% & 0.00\% & 0.00\% & 0.00\% & 0.00\% & 0.00\% & 0.00\% & - \\
& summaries + CoT & 0.00\% & 0.00\% & 0.00\% & 0.00\% & 0.26\% & 0.00\% & 0.00\% & 0.00\% & - \\ \hline

\multirow{6}{*}{} &  & 1,2,3,4 & 5,6,7,8 & 10,11,12,13 & 15,16,17,18 & 17,18,19,20 & 1,3,5,7 & 5,7,9,11 & 10,12,14,16 & 14,16,18,20 \\ \hline
\multirow{6}{*}{4-hop} & standard & 0.00\% & 0.00\% & 0.00\% & 0.00\% & 0.00\% & 0.00\% & 0.00\% & 0.00\% & 0.00\% \\
& standard + CoT & 0.00\% & 0.00\% & 0.00\% & 0.00\% & 0.00\% & 0.00\% & 0.00\% & 0.00\% & 0.00\% \\
& kg & 0.00\% & 0.00\% & 0.00\% & 0.00\% & 0.00\% & 0.00\% & 0.00\% & 0.00\% & 0.00\% \\
& kg + CoT & 0.00\% & 0.00\% & 0.00\% & 0.00\% & 0.00\% & 0.00\% & 0.00\% & 0.00\% & 0.00\% \\
& summaries & 0.00\% & 0.00\% & 0.00\% & 0.00\% & 0.00\% & 0.00\% & 0.00\% & 0.00\% & 0.00\% \\
& summaries + CoT & 0.00\% & 0.00\% & 0.00\% & 0.00\% & 0.00\% & 0.00\% & 0.00\% & 0.00\% & 0.00\% \\ \hline

\multirow{6}{*}{}2wiki (31.65\%) &  & 1,2 & 5,6 & 10,11 & 15,16 & 19,20 & 1,5 & 5,10 & 10,15 & 15,20 \\ \hline
\multirow{6}{*}{2-hop} & standard & 27.29\% & 5.02\% & 3.15\% & 3.09\% & 2.54\% & 17.39\% & 4.65\% & 3.17\% & 2.82\% \\
& standard + CoT & 0.26\% & 0.10\% & 0.06\% & 0.24\% & 0.16\% & 0.22\% & 0.26\% & 0.16\% & 0.14\% \\
& kg & 25.58\% & 7.90\% & 6.32\% & 6.44\% & 6.20\% & 17.53\% & 7.86\% & 6.52\% & 6.12\% \\
& kg + CoT & 0.81\% & 0.77\% & 0.89\% & 1.00\% & 1.02\% & 1.00\% & 1.06\% & 1.02\% & 1.02\% \\
& summaries & 28.99\% & 12.37\% & 4.31\% & 4.06\% & 3.51\% & 20.85\% & 10.20\% & 4.02\% & 3.82\% \\
& summaries + CoT & 0.12\% & 0.10\% & 0.08\% & 0.06\% & 0.06\% & 0.06\% & 0.10\% & 0.10\% & 0.08\% \\ \hline

\multirow{6}{*}{} &  & 1,2,3,4 & 5,6,7,8 & 10,11,12,13 & 15,16,17,18 & 17,18,19,20 & 1,3,5,7 & 5,7,9,11 & 10,12,14,16 & 14,16,18,20 \\ \hline
\multirow{6}{*}{4-hop} & standard & 30.75\% & 8.35\% & 5.42\% & 4.54\% & 3.95\% & 24.30\% & 8.64\% & 5.34\% & 3.73\% \\
& standard + CoT & 0.51\% & 0.51\% & 0.51\% & 1.10\% & 0.73\% & 0.51\% & 0.51\% & 0.88\% & 0.73\% \\
& kg & 43.56\% & 19.25\% & 14.49\% & 14.71\% & 14.49\% & 34.99\% & 19.33\% & 14.20\% & 13.69\% \\
& kg + CoT & 2.56\% & 3.15\% & 2.12\% & 2.34\% & 2.34\% & 1.98\% & 2.64\% & 2.27\% & 2.42\% \\
& summaries & 56.95\% & 16.91\% & 5.93\% & 5.20\% & 4.03\% & 36.53\% & 14.86\% & 5.71\% & 4.47\% \\
& summaries + CoT & 0.12\% & 0.00\% & 0.00\% & 0.07\% & 0.07\% & 0.06\% & 0.00\% & 0.00\% & 0.00\% \\ \hline

\end{tabular}
\caption{Full experimental results of llama-2-7b-longlora-8k-ft. Percentages next to dataset names are the closed-book scores for the full set.}
\label{table:performance}
\end{table*}

\end{document}